\documentclass[runningheads]{llncs}
\usepackage{graphicx}
\usepackage{amsmath}
\usepackage{amssymb}
\usepackage{bm}
\usepackage{makecell}
\usepackage{booktabs}
\usepackage{mwe}
\usepackage{color}
\usepackage{hyperref}
\hypersetup{
    colorlinks = true,
    linkcolor = blue,
    anchorcolor = blue,
    citecolor = blue,
    filecolor = blue,
    urlcolor = blue
}

\newcommand{\rln}{\mathbb{R}}
\newcommand{\norm}[1]{\left\lVert#1\right\rVert}
\newcommand{\tmplt}{\mathcal{T}}

\begin{document}
\title{Landmark-free Statistical Shape Modeling via Neural Flow Deformations}

\newcommand{\repeatthanks}{\textsuperscript{\thefootnote}}

\author{David L\"udke\thanks{Equal contribution}\inst{1}
\and
Tamaz Amiranashvili\repeatthanks\inst{1,2,3} %\orcidID{0000-0001-8914-3427}
\and
Felix Ambellan\inst{1,4} %\orcidID{0000-0001-9415-0859}
\and
Ivan Ezhov\inst{3}
\and
Bjoern Menze\inst{2,3}%\orcidID{0000-0003-4136-5690}
\and
Stefan Zachow\inst{1}%\orcidID{0000-0001-7964-3049}}
}
% index{L\"udke, David}
% index{Amiranashvili, Tamaz}
% index{Ambellan, Felix}
% index{Ezhov, Ivan}
% index{Menze, Bjoern}
% index{Zachow, Stefan}

\authorrunning{D. L\"udke and T. Amiranashvili et al.}

\institute{Zuse Institute Berlin, Berlin, Germany \and
University of Zurich, Zurich, Switzerland \\
\email{tamaz.amiranashvili@uzh.ch} \and
Technical University of Munich, Munich, Germany \and
Freie Universit\"at Berlin, Berlin, Germany \\
}

\maketitle

\begin{abstract}

Statistical shape modeling aims at capturing shape variations of an anatomical structure that occur within a given population.
Shape models are employed in many tasks, such as shape reconstruction and image segmentation, but also shape generation and classification.
Existing shape priors either require dense correspondence between training examples or lack robustness and topological guarantees.
We present FlowSSM, a novel shape modeling approach that learns shape variability without requiring dense correspondence between training instances.
It relies on a hierarchy of continuous deformation flows, which are parametrized by a neural network.
Our model outperforms state-of-the-art methods in providing an expressive and robust shape prior for distal femur and liver.
We show that the emerging latent representation is discriminative by separating healthy from pathological shapes.
Ultimately, we demonstrate its effectiveness on two shape reconstruction tasks from partial data.
Our source code is publicly available\footnote{\url{https://github.com/davecasp/flowssm}}.

\end{abstract}

\section{Introduction}
Statistical shape models (SSMs) are an important tool in medical image analysis and computational anatomy.
Application examples include diagnosis, pathology detection and segmentation, outlier detection, shape reconstruction, etc. (cf.\ e.g.\ \cite{ambellan2019statistical}).
For this purpose, shape models capture variations in shape that occur within the population of a given anatomical structure.
The model is typically learned in an automated fashion from example data.
This results in a generative model, which can be, among others, used as a regularizer for ill-posed tasks.
Furthermore, shape models provide a low-dimensional statistics-driven latent representation of shapes, which allows clustering, classification and shape regression.

The most established statistical shape model in the medical domain is a \textit{point distribution model} (PDM) \cite{cootes1995active,heimann2009statistical}.
While PDMs are simple and effective, they employ linear statistics, which restricts their capacity to fully capture  the high variability of anatomical structures \cite{davis2010population}.
There exists a variety of non-linear statistical shape models \cite{fletcher2004principal,huckemann2010intrinsic,von2018efficient} that have been shown to increase the flexibility and can handle large, non-linear deformations well, such as the recently developed \textit{fundamental coordinate model} (FCM) \cite{ambellan2021rigid}.
However, most of these models still require consistently parametrized training shapes, i.e.\ shapes which are in dense correspondence to each other.
This limits the learned shape model to the resolution of the given training data.
Furthermore, generating well-defined dense correspondence typically requires manual landmark annotation that is tedious to obtain.
Moreover,  establishing  exact  dense  correspondence  is  faulted  by  generally  undefined true dense correspondence of biological shapes \cite{heimann2009statistical}.
Lastly, as these models work on uncoupled primitives, such as mesh triangles, they have no inherent safeguards to avoid the generation of unnatural shapes with self-intersections.

Another popular tool in computational anatomy is \textit{large deformation diffeomorphic metric mapping} (LDDMM) \cite{miller2006geodesic,vaillant2007diffeomorphic,durrleman2014morphometry}.
It allows estimating a template from a population of training shapes, as well as deformations that map the template to individual subjects.
These deformations are continuous, however they are parametrized by a finite set of momentum vectors, providing a compact latent representation.
While the template is population-wide, the subject-specific latent representations result from independent, pair-wise registration of each subject to the template.
In context of shape modeling, LDDMMs have several advantages over PDMs and FCMs.
First, LDDMMs do not rely on dense correspondence between training shapes.
Second, being formulated as a diffeomorphism, they are less prone to producing self-intersections.
Recently, novel parametrizations of approximately diffeomorphic deformation flows through a multilayer perceptron (MLP) were proposed \cite{niemeyer2019occupancy,jiang2020shapeflow,gupta2020neural}.
The MLP's parameters are estimated on the training data, hence representing a population-wide prior.
In this work, we extend neurally parametrized deformation fields to create \emph{FlowSSM} -- a novel hierarchical, continuous shape model, that is able to accurately capture variations in shapes of anatomical structures without requiring dense correspondence between training shapes.
Our main contributions can be summarized as follows:
\begin{itemize}
    \item We propose a novel shape model FlowSSM, which is based on continuous neural flows and produces natural shape deformations without relying on a given dense correspondence between training shapes.
    \item Our model outperforms state-of-the-art methods in capturing variability in anatomical shapes variation on liver and distal femur, as measured in generalization ability and specificity.
    \item We demonstrate the discriminative nature of our learned latent representation by classifying knee osteoarthritis.
    \item We qualitatively show that our model provides anatomically plausible reconstructions from partial and sparse shapes on both anatomical structures.
\end{itemize}

\begin{figure}[t]
    \centering
    \includegraphics[width=\textwidth]{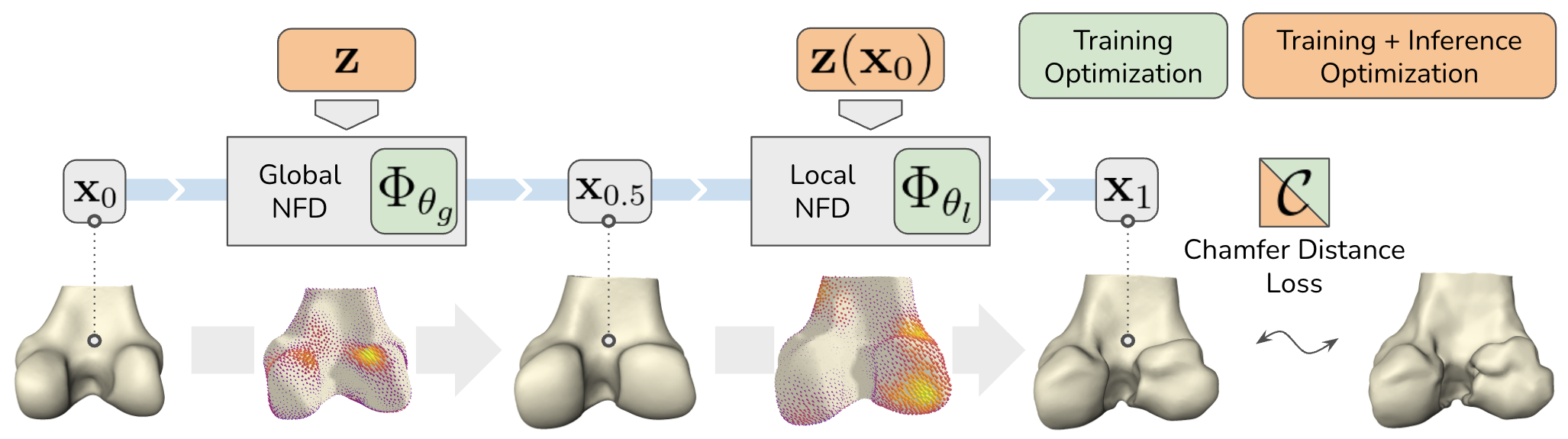}
    \caption{We model shapes by template deformations, defined by two sequential, continuous neural flow deformers (NFD). The model is trained as an auto-decoder with correspondence-free Chamfer distance loss.
    The bottom row visualizes the template deformation through the velocity fields.
    }
    \label{fig:method}
\end{figure}

\section{Methodology}

\subsection{Neural Flow Deformer}
We model shapes by a continuous deformation of a template surface (cf.\ Fig.\ \ref{fig:method}).
A shape $X$ is obtained by deforming each point $\bm{x}_0$ on a template $\tmplt$ along a trajectory $\bm{x}: [0, 1] \to \rln^{3}$, where $\bm{x}(0) = \bm{x}_0$ and $\bm{x}(1) \in X$.
We model these trajectories by a time-dependent velocity field $\bm{v}: \rln^{3} \times [0,1] \to \rln^{3}$, that relates to $\bm{x}(\cdot)$ through the following ordinary differential equation:
\begin{align}
    \frac{\mathrm{d} \bm{x}(t)}{\mathrm{d}t} = \bm{v}(\bm{x}(t), t).
\end{align}
For each initial condition $\bm{x}_0 \in \tmplt$, the solution of this ODE describes a trajectory that starts in $\bm{x}_0$.
All solutions for different starting points are summarized as the flow function $\Phi: \rln^3 \times [0, 1] \to \rln^3$, for which holds:
\begin{align}
\Phi(\bm{x}_0, \tau) = \bm{x}_0 + \int_0^{\tau}\bm{v}(\Phi(\bm{x}_0, t), t)\,dt.
\label{eq:ode}
\end{align}
Hence, the flow $\Phi$ describes the trajectory of a point $\bm{x}_0$ which tangentially follows the velocity field $\bm{v}$.
The target shape is described by $\{\Phi(\bm{x}_0, 1) \mid \bm{x}_0 \in \tmplt\}$.

\subsubsection{Velocity Field Parametrization}
Following recent advances in modeling continuous neural flows \cite{chen2018neuralode,niemeyer2019occupancy,jiang2020shapeflow}, we propose to parametrize the velocity field $\bm{v}(\cdot,\cdot)$ through an MLP that is conditional on a shape-specific latent representation $\bm{z} \in \rln^d$.
Similar to \cite{jiang2020shapeflow}, we define the deformation velocity field as:
\begin{align}
    \bm{v}_{\theta}(\bm{x}(t),t)=\bm{\mathnormal{f}}_{\theta}(\bm{x}(t),t\cdot\bm{z})\cdot\norm{\bm{z}}_{2},
\label{eq:velocity_param}
\end{align}
where $\bm{\mathnormal{f}}$ is the flow function parametrized by an MLP (IM-Net \cite{chen2019learning}) and then scaled by a flow magnitude proportional to the euclidean vector norm of $\bm{z}$. For architectural detail of the applied flow function, we refer the reader to Fig.\ \ref{fig:backbone} in the supplements.

\subsubsection{Global and Local Level of Detail}
In Eq. \eqref{eq:velocity_param}, one global latent vector $\bm{z} \in \rln^d$ for all starting points is employed.
While being compact, such global parametrization tends to produce smooth, low-frequency deformations \cite{jiang2020shapeflow}.
To extend the deformations' frequency spectrum, we propose to model the latent representation as a continuous function in $\rln^3$.
In particular, we interpolate $M$ latent weights $\bm{z}_k \in \rln^d$ at their control point positions $\bm{c}_k \in \rln^3$ via radial basis functions, using a Gaussian kernel with an inverse width $\varepsilon_k$:
\begin{equation}
\begin{split}
    \bm{z}(\bm{x}) &= \sum \limits _{k=1}^{M}\bm{z}_k \varphi_k(\norm{\bm{c}_k-\bm{x}}_2)\,, \\
    \varphi_k(r) &= e^{-(\varepsilon_k \cdot r)^2}\,.
\end{split}
\label{eq:rbf}
\end{equation}
When solving \eqref{eq:ode} to obtain a deformation trajectory for a starting point $\bm{x}_0 \in \tmplt$, we compute $\bm{z}(x)$ via \eqref{eq:rbf} only at the starting point $\bm{x}_0$, using a constant $\bm{z}(\bm{x}_0)$ for the whole resulting trajectory.
This allows us to sample $\{\bm{c}_k\}$ equidistantly on the template surface, instead of the whole volume, making for a more compact parametrization.

To disentangle global, low frequency shape characteristics from local, high frequency details, we apply two deformers sequentially.
The first one uses a global latent vector, while the second one uses the localized formulation \eqref{eq:rbf}.
Each deformer has its own set of MLP parameters $\theta_g$ and $\theta_l$ respectively.

In summary, a global latent vector $\bm{z}$, $M$ local control points $(\bm{z}_k, \bm{c}_k, \varepsilon_k)$ and MLP parameters $\theta_g$ and $\theta_l$ parametrize a continuous, time-dependent velocity field, which defines deformation trajectories for every surface point on a template, yielding a single target shape.

\subsection{Training}
Given a set of $N$ arbitrarily parametrized but topologically equivalent training surfaces $\mathcal{X}=\{X_1,\cdots,X_N\}$ and a template surface $\mathcal{T}$, our model is trained without given dense correspondence.
Every shape $X_i$ is represented by $M + 1$ latent vectors $(\bm{z}^i, \{\bm{z}_k^i\})$.
The MLP parameters, $\{\bm{c}_k\}$, $\varepsilon_k$ and $\mathcal{T}$ are shared across all training surfaces, representing a population-wide prior.
We train our model as an auto-decoder \cite{park2019deepsdf,jiang2020shapeflow}.
That is, in training, the latent representations are randomly initialized from $\mathcal{N}(0, 0.1^2)$ and jointly optimized with the MLP parameters, without learning an encoder.
The global and local deformers are trained consecutively with the same target surfaces $\mathcal{X}$.
The loss is the correspondence-free, symmetric point-set to point-set Chamfer distance $\mathcal{C}$ between randomly sampled surface points of the target $P_i\subset X_i$ and deformed, sampled surface points of the template $P_\Phi=\Phi^i(P_{\tmplt} \subset \tmplt, 1)$:
\begin{align}
    \mathcal{C}(P_i,P_\Phi) = \frac{1}{2|P_i|}\sum \limits _{\bm{x}_i\in P_i}\min_{\bm{x} \in P_\Phi} \norm{\bm{x}_i-\bm{x}}_2 + \frac{1}{2|P_\Phi|}\sum \limits _{\bm{x}\in P_\Phi}\min_{\bm{x}_i \in P_i} \norm{\bm{x}_i-\bm{x}}_2\,.
    \label{eq:loss}
\end{align}

After training, we estimate the probability distribution of the latent vectors by performing principal component analysis (PCA) \cite{wold1987principal} on the emerging latent representations of training shapes.
The PCA is performed separately for the global and local latent representations, keeping all modes of variation.

\subsection{Inference}
To embed unseen shape instances in the learned shape space, e.g.\ for reconstruction, one needs to obtain a latent representation that best describes a given observation.
For this, a new latent embedding is initialized from $\mathcal{N}(0, 0.1^2)$.
We fix the trained MLP parameters and $\{\varepsilon_k\}$, while the latent representations are optimized with the correspondence-free loss \eqref{eq:loss}.
Within this process, the latent embedding is restricted to the subspace, i.e.\ span of the training representations, defined by the PCA.
The global and local latent representations are trained consecutively.

\section{Experiments}

We conduct experiments on three datasets: liver, distal femur bone, as well as the condyle region of the distal femur (a dataset summary can be found in Table \ref{table:datasets_details} in the supplements).
Both femur datasets cover the full spectrum of osteoarthritis severity.
The liver, being a deformable soft tissue organ, exhibits a much higher degree of nonlinear variability, yielding a particularly challenging dataset for shape modeling.

\textbf{Liver and Femur Datasets} The datasets consist of correspondingly 112 and 253 surfaces.
The surfaces were obtained from manually segmented CT scans of liver \cite{kainmuller2007shape} and MRI scans of knee \cite{ambellan2019automated}.
For correspondence-based methods, consistently meshed surfaces were obtained semi-automatically following \cite{kainmuller2007shape,von2018efficient}.
Both datasets are split into 70\% training, 10\% validation, and 20\% test data.

\textbf{Classification Dataset} We adopt the \emph{Osteoarthritis Initiative\footnote{nda.nih.gov/oai} - right distal femur} dataset from \cite{von2018efficient,ambellan2021rigid}.
The dataset consists of 116 femur surfaces with 58 healthy cases and 58 severe cases of osteoarthritis, according to their Kellgren and Lawrence \cite{kellgren1957radiological} score of 0/1 and 4, respectively.
For further information on the dataset and its construction, cf.\ \cite{von2018efficient,ambellan2021rigid}.

\textbf{Template Generation} For liver and femur datasets, we generate a correspondence free template per dataset by applying the hub-and-spokes approach of ShapeFlow \cite{jiang2020shapeflow}.
This results in non-rigidly aligned training shapes in a canonical space.
These are used for the template surface generation by applying an unscreened Poisson surface reconstruction \cite{kazhdan2013screened}.
Finally, the template meshes are simplified to attain approximately the same number of faces as in the training data.
For the classification dataset, the template is directly constructed by unscreened Poisson surface reconstruction from given shapes.

\textbf{Data Preprocessing}
Although our method randomly samples points on training surfaces and therefore does not use correspondences, we further emphasize this fact by re-meshing the surfaces to destroy correspondences.
Each mesh is aligned to its template via Iterative Closest Points \cite{besl1992method}. 
All meshes are then isotropically rescaled to fit $[-1, 1]$ based on the largest training instance and centred around the origin in alignment with the template shape.
The classification dataset is left unchanged for comparability to previous works \cite{von2018efficient,ambellan2021rigid}.

\begin{table}
\centering
\caption{Generality (measured as average symmetric surface distance in mm), specificity (measured as average Chamfer distance in mm) and number of self-intersecting meshes (SIM) for distal femur and liver.
Bold numbers indicate statistically significant improvements to baselines (paired t-test, $p < 0.05$).}
\setlength{\tabcolsep}{8pt}
\begin{tabular}{c c c c c}

\toprule
\rule{0pt}{0pt}
\emph{\textbf{Femur}} &
\makecell{Generality \\{} [mm] $\downarrow$ } &
\makecell{Specificity \\{} [mm] $\downarrow$ } &
\makecell{SIM Gen.\\{} [\#] $\downarrow$} &
\makecell{SIM Spec.\\{} [\#] $\downarrow$} \\
\hline
PDM  & 0.26 $\pm$ 0.04 & 1.12 $\pm$ 0.12 & 1 & 25\\ %final
FCM  & 0.26 $\pm$ 0.04 & 1.15 $\pm$ 0.16 & 0 & 1 \\
\hline
LDDMM  & 0.27 $\pm$ 0.04 & 1.19 $\pm$ 0.12 & 0 & 0 \\ %final
FlowSSM (ours)   & \textbf{0.24} $\pm$ \textbf{0.04} & \textbf{0.97} $\pm$ \textbf{0.09} & 0 & 0  \\ %final
\hline
\hline
\rule{0pt}{0pt} \thead{\emph{\textbf{Liver}}\\} & & & &  \\
\hline
PDM  &  1.57 $\pm$ 0.30 & 4.98 $\pm$ 0.59  & 18  & 871\\ %final
FCM  & 1.52 $\pm$ 0.26 &  5.91 $\pm$ 0.90& 7 & 668 \\ %final
\hline
LDDMM  &  $1.75 \pm 0.36$ & $ 5.16 \pm 0.60 $& 4 & 86\\ %final
FlowSSM (ours)  & \textbf{1.20} $\pm$ \textbf{0.18} & \textbf{4.00} $\pm$ \textbf{0.48}  & 5 & 4 \\ %final

\bottomrule

\end{tabular}

\label{table:generality_specificity}

\end{table}

\subsection{Experimental Setup}

In the following, we compare FlowSSM to two correspondence-based models -- the linear PDM \cite{cootes1995active} and a non-linear statistical shape model, the FCM \cite{ambellan2021rigid,AmbellanHanikvonTycowicz2021Morphom}.
Furthermore, we compare to an LDDMM-based approach \cite{durrleman2014morphometry}, where the emerging latent shape space is estimated by a PCA on the learned initial momentum vectors of training shapes.
This is motivated by \cite{wang2007large,charon2021landmark}, who used a PCA on momenta for disease classification.
For comparability, we use the same template and control points as in our method.
We emphasize that the PDM and FCM baselines require corresponding meshes for training, in contrast to LDDMM and ours.
We run all experiments on an NVIDIA A100 40GB GPU.
The training takes about 2h and 4h for liver and femur data, respectively, while generating a single shape given a latent representation takes about 0.14s per shape.
For further implementation details and hyperparameters, refer to Tables \ref{table:hyperparameters} and \ref{tab:software_versions} in the supplements.

\subsection{Generality and Specificity}
\label{sec:gen_spec}
To quantitatively assess the quality of our generative model in the light of statistical shape modeling, we apply two standard measures, generalization ability and specificity, as introduced by Davies \cite{davies2002learning}. 
Here, generality refers to the ability of a statistical shape model to fit unseen instances of the class of shapes.
It is evaluated by the average symmetric surface distance between all test shapes and their reconstructions.
To ensure comparability between the methods, we fit the correspondence-based models to the test shape by minimizing the average surface distance.
In contrast to generality, specificity measures the restrictiveness of a model to generate shapes that are close to training instances.
To generate random samples, we utilize the distributions described by all PCA modes of each model.
We randomly sample 1000 surfaces by sampling latent representations and computing the surfaces. We then calculate their average surface distance to the closest example in the training set.
For computational reasons, the surface distance is approximated by symmetric Chamfer distance.

The results in Table \ref{table:generality_specificity} show that FlowSSM exhibits the lowest generality error for both liver and femur, i.e., best fit to unseen shape instances.
Further, FlowSSM is more specific than the baselines and generates samples closest to the training population.
All methods are less precise on the liver dataset, which can be explained by larger, nonlinear variability in the deformable liver.
Generally, dense correspondence between livers is not clearly defined and thus poses an additional challenge for PDM and FCM.
The femur, on the other hand, has better defined correspondences and less variation between shapes, which we conjecture to be the main reason for similar performance between all methods.
Furthermore, self-intersections degrade the resulting mesh quality and suggest that a model produces unnatural shape deformations.
FlowSSM exhibits by far the lowest total number of self-intersecting meshes out of the four models, indicating smooth and natural shape deformations.

Lastly, our localized deformer allows us to nearly half the generalization error, when compared to the global parametrization of \cite{jiang2020shapeflow,niemeyer2019occupancy} (cf.\ Table \ref{table:gen_ablation} in the supplements).

\begin{figure}
\centering
    \begin{minipage}{.48\linewidth}
        \centering
        \includegraphics[width=1.0\textwidth]{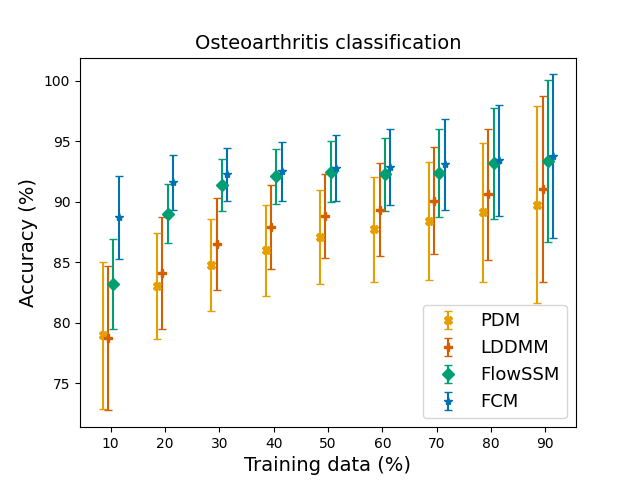}
    \end{minipage}
    ~ 
    \begin{minipage}{.48\textwidth}
        \centering
        \includegraphics[width=1.0\textwidth]{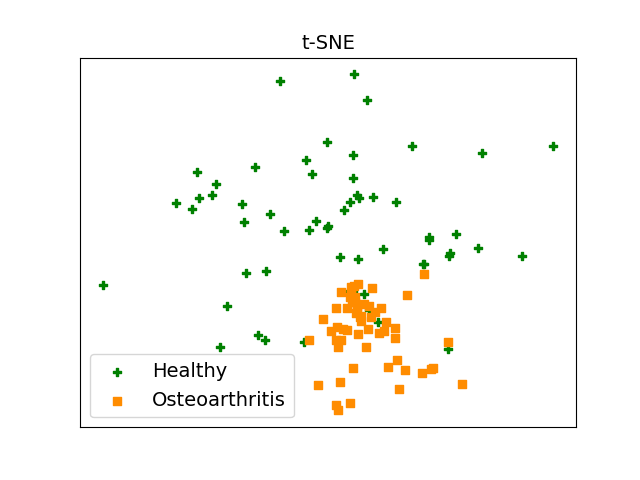}
    \end{minipage}
    \caption{Left: Osteoarthritis classification accuracy indicated by mean and standard deviation of the 10,000 balanced split samples per partitioning. Right: Unsupervised 2D t-SNE visualization of the learned PCA weights of FlowSSM shows a clustering of diseased cases.}

\label{fig:classification}
\end{figure}

\subsection{Osteoarthritis Classification}

A discriminative emerging latent space is an important property of a shape model.
Following the experimental setup of \cite{ambellan2021rigid,von2018efficient}, we evaluate our model's ability to classify knee osteoarthritis (OA) -- a degenerative disease of the joints which causes pathological malformities of the femur.
To this end, FlowSSM is trained to best reconstruct the complete classification data of the condyle region.
Note that this does not involve the OA labels and is therefore to be considered unsupervised.
Ultimately, we classify the respective PCA weights of our model and the baselines with a linear support vector machine.
We evaluate the average classification accuracy of a stratified Monte-Carlo cross-validation for 10,000 samples per partitioning with training set sizes from 10\% to 90\%.

In Fig.\ \ref{fig:classification}, the high average classification accuracy of our model across all partitionings is indicative of the discriminative nature of the unsupervisedly learned latent space.
This is further exemplified through the emerging clustering in the t-distributed Stochastic Neighbor Embedding (t-SNE) \cite{van2008visualizing} visualization.
The FCM provides slightly more robust classification, especially for smaller training sets, indicating a more distinct linear separability of the weights.
Still, our model shows comparable performance for larger training sets, without relying on given correspondence in representation learning.

\subsection{Shape Reconstruction}

\begin{figure}
\centering
\includegraphics[width=1.0\textwidth]{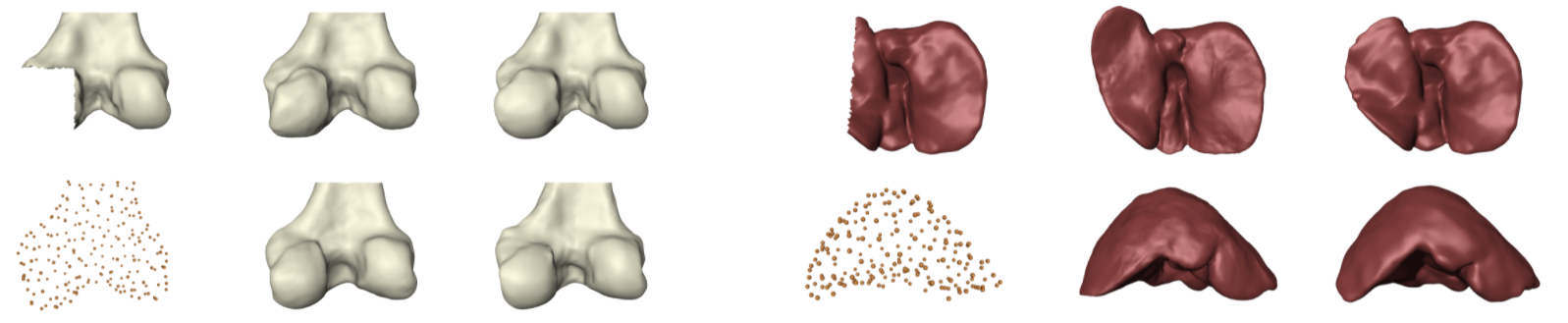}
\caption{Femur and liver reconstructions (center) from sparse point-sets and partial meshes (left) with their respective ground truth (right).}
\label{fig:reconstruction}
\end{figure}

Our model can be fitted to various target parametrizations.
We fit our trained model from Sec.\ \ref{sec:gen_spec} to a partial mesh and a sparse point-cloud (200 surface points) from unseen subjects and reconstruct full meshes.
In contrast to the shape inference in the generality experiment, a one-sided Chamfer distance is minimized to obtain a latent embedding.
Reconstructions in Fig.\ \ref{fig:reconstruction} qualitatively highlight our model's specificity by displaying anatomically plausible reconstructions.
At the same time, our model's generality is demonstrated by preserving patient-specific details, present in the targets.

\section{Conclusion}
We present a novel landmark-free shape model that captures variability of a given population of anatomical shapes.
This is achieved through a novel, localized neural flow parametrization.
Our model exhibits good generalization ability to unseen shapes without sacrificing specificity, demonstrated on two anatomical structures.
Additionally, in contrast to established methods, our model does not require any given dense correspondences.
Furthermore, the emerging latent space is discriminative, as demonstrated by classification of osteoarthritis.

In practice, our method simplifies building robust and expressive shape priors by utilizing large pools of heterogeneous training data, independent of their parametrization and resolution.
The emerging latent representation is useful in discriminating various shape-correlated features.
As the flow parametrization of our model makes it well suited for studying temporal shape trajectories, this poses an exciting future research direction.

\section*{Acknowledgements}
This work was supported by the Bundesministerium f\"ur Bildung und Forschung (BMBF) through the research campus MODAL (ref. 3FO18501) and The Berlin Institute for the Foundations of Learning and Data (BIFOLD) - (ref. 01IS18025A and ref. 01IS18037A). We are grateful for the open-access dataset OAI.
(The Osteoarthritis Initiative is a public-private partnership comprised of five contracts 
(N01-AR-2-2258; N01-AR-2-2259; N01-AR-2-2260; N01-AR-2-2261; N01-AR-2-2262) 
funded by the National Institutes of Health, a branch of the Department of Health and 
Human Services, and conducted by the OAI Study Investigators. Private funding partners 
include Merck Research Laboratories; Novartis Pharmaceuticals Corporation, 
GlaxoSmithKline; and Pfizer, Inc. Private sector funding for the OAI is managed by the 
Foundation for the National Institutes of Health. This manuscript was prepared using an OAI 
public use data set and does not necessarily reflect the opinions or views of the OAI 
investigators, the NIH, or the private funding partners.)

\bibliographystyle{splncs04}
\bibliography{bib}
\newpage
\appendix

\renewcommand{\thefigure}{S\arabic{figure}}
\setcounter{figure}{0}
\renewcommand{\thetable}{S\arabic{table}}
\setcounter{table}{0}

%\pagenumbering{arabic}
\section{Supplements}

\begin{figure}
    \centering
    \includegraphics[width=0.95\textwidth]{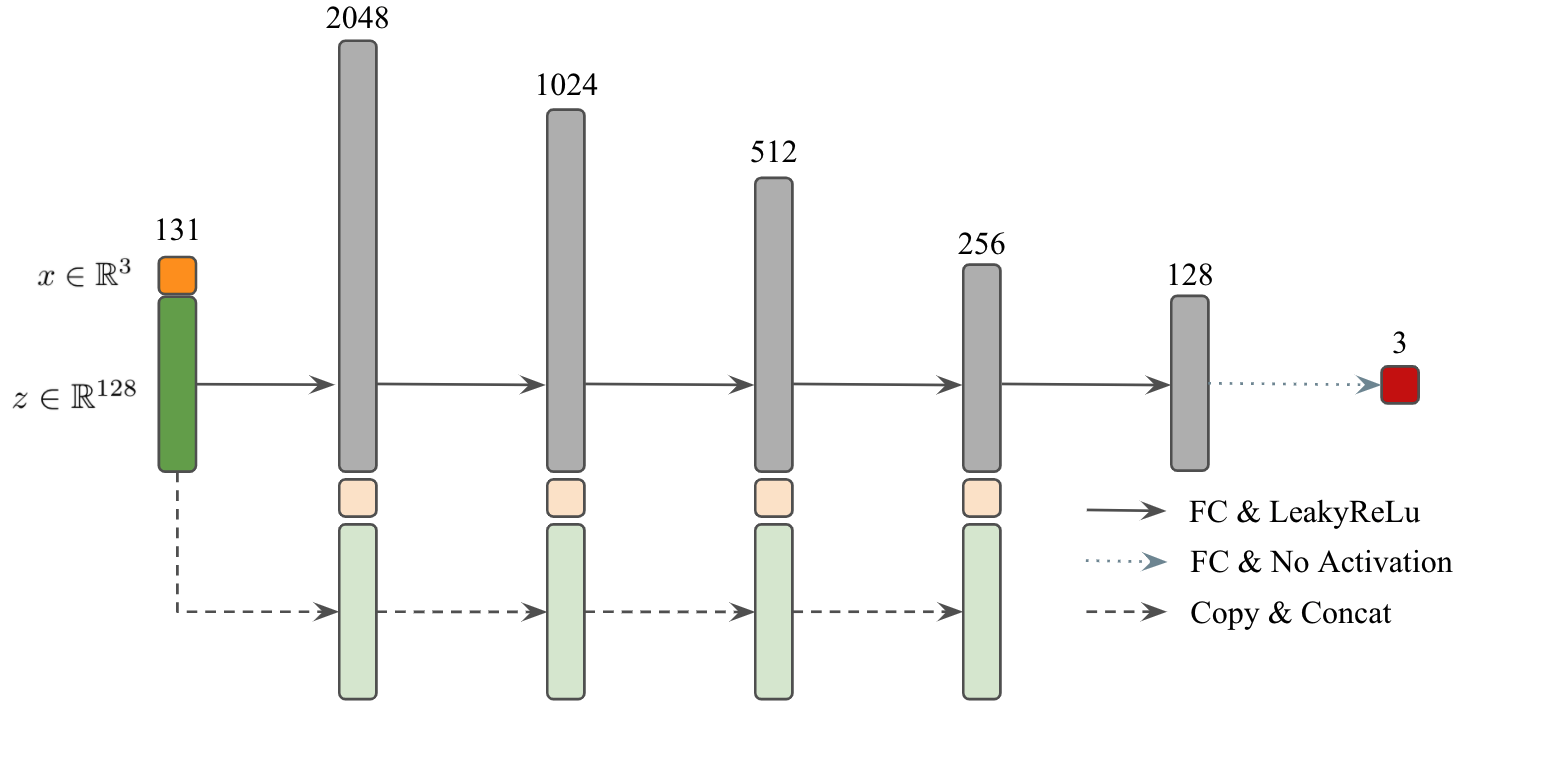}
    \caption{Depiction of the IM-Net MLP network \cite{chen2019learning} architecture of the neural flow field deformer. The IM-Net network consists of 5 fully connected layers with intermediate LeakyReLU activations and skip-layer connections (copy and concatenate). It takes a point coordinate $x$ and a latent vector $z$ as an input and outputs a displacement vector.}
\label{fig:backbone}
\end{figure}

\begin{table}
\centering
\caption{
Dataset summary of the number of shapes in the training, validation and test set and number of vertices and faces per shape.
\textit{Note}: In the classification task, FlowSSM was trained on the whole dataset to best reconstruct the training shapes.
This relates to an unsupervised learning task, as the classification labels were not used in training.
}

\setlength{\tabcolsep}{12pt}
\begin{tabular}{c  c  c  c }

\toprule
 & Liver & Femur & Classification \\
\hline
\# Shapes  & 112 & 253 & 116  \\ %final
\# Vertices per shape  & 12,974 & 11,968 & 8,988   \\ %final
\# Faces per shape  & 25,944 & 23,814 & 17,829  \\ %final
\# Training shapes  & 78 & 177 & 116  \\ %final
\# Validation shapes  & 11 & 25 & --  \\ %final
\# Test shapes  & 23 & 51 & --  \\ %final
\bottomrule
\end{tabular}
\label{table:datasets_details}
\end{table}

\begin{table}
\centering
\caption{Hyperparameter values for different data sets. The number of control points $M \in [4, 10]$ and $\varepsilon \in [0.01, 30]$ of the radial basis function interpolation were tuned with Optuna TPESampler (optuna.readthedocs.io). The hyperparameter search minimizes the generalization ability evaluated on the validation data for femur and liver and a training reconstruction error for the classification data.
$lr$ refers to the Adam optimizer learning rate and $d$ is the dimensionality of the latent representation.
For LDDMM baselines, we use recommended parameters from the open-source reference implementation Deformetrica.
}

\setlength{\tabcolsep}{12pt}
\begin{tabular}{c  c  c  c }

\toprule
Parameter & Liver & Femur & Classification \\
\hline
\rule{0pt}{10pt}
$M$  & $6^3$ & $7^3$ &  $10^3$ \\ %final
initial $\varepsilon_k$  & 2.5963 & 3.7606 & 14.3416  \\ %final
train $lr$  & 0.001 & 0.001 & 0.001  \\ %final
train epochs  & 300 & 300 & 300  \\ %final
inference $lr$  & 0.01 & 0.01 & 0.01  \\ %final
inference epochs  & 600 & 600 & 600  \\ %final
batch size  & 16 & 16 & 16 \\ %final
\# sampled surface points  & 15,000 & 15,000 & 15,000 \\ %final
$d$  & 128 & 128 & 128 \\ %final
\bottomrule

\end{tabular}
\label{table:hyperparameters}

\end{table}

\begin{table}
\centering
\caption{Ablation study on the average surface distance in the generality experiment of the global vs. global + local flow deformer. Note, that the parametrization of the global flow deformer corresponds to the parameterization applied in ShapeFlow \cite{jiang2020shapeflow}, which only insufficiently represents the geometric details present in anatomical structures.}

\setlength{\tabcolsep}{12pt}
\begin{tabular}{c  c  c }

\toprule
 & Liver & Femur \\
\hline
Global Flow Deformer  & 1.96 $\pm$ 0.43 & 0.41 $\pm$ 0.08   \\ %final
Global + Local Flow Deformer  & 1.20 $\pm$ 0.18 &  0.24  $\pm$ 0.04  \\ %final
\bottomrule

\end{tabular}
\label{table:gen_ablation}

\end{table}

\begin{table}
\centering
\caption{Versions of used software packages.}
\setlength{\tabcolsep}{12pt}
\begin{tabular}{c  c}
\toprule
Package & Version \\
\hline
PyTorch (pytorch.org) & 1.10.0 \\
torchdiffeq (github.com/rtqichen/torchdiffeq) & 0.2.2 \\
Morphomatics (morphomatics.github.io) & 1.0 (curious caesar) \\
Deformetrica (deformetrica.org) & 4.3.0 \\

\bottomrule

\end{tabular}
\label{tab:software_versions}
\end{table}

\end{document}